\pdfoutput=1

\documentclass[11pt]{article}

\usepackage{acl}
\usepackage{times}
\usepackage{latexsym}
\usepackage{amsmath}
\usepackage{booktabs}
\usepackage{makecell}
\usepackage{pifont} 
\newcommand{\cmark}{\ding{51}} 
\newcommand{\xmark}{\ding{55}} 
\usepackage{amssymb}
\usepackage{float}

\usepackage[T1]{fontenc}

\usepackage[utf8]{inputenc}

\usepackage{microtype}

\usepackage{inconsolata}

\usepackage{graphicx}

\usepackage{ragged2e}
\usepackage{multirow}
\usepackage{graphicx}
\usepackage{xcolor}
\usepackage{xspace}
\usepackage{color}
\usepackage{colortbl}
\usepackage{tablefootnote}
\usepackage{pifont}
\usepackage{makecell}
\usepackage[most]{tcolorbox}
\usepackage{framed}
\usepackage{mdframed}
\usepackage{caption}
\usepackage{subcaption}
\usepackage{longtable}
\usepackage{float}
\usepackage{booktabs}
\usepackage{arydshln} 

\usepackage{amsfonts}
\usepackage{booktabs}

\newcommand{\paratitle}[1]{\vspace{1.5ex}\noindent\textbf{#1}}

\newcommand{\ie}{\emph{i.e.,}\xspace}

\newcommand{\eg}{\emph{e.g.,}\xspace}

\newcommand{\ignore}[1]{}

\usepackage{algpseudocode}
\usepackage{amsthm}
\usepackage{amsmath}
\usepackage{tikz}
\usepackage{amsmath}

\definecolor{mygray}{gray}{0.97}
\colorlet{shadecolor}{mygray}

\newmdenv[%
  backgroundcolor=mygray, 
  linewidth=0pt
]{newshaded}

\title{LinkAlign: Scalable Schema Linking for Real-World Large-Scale Multi-Database Text-to-SQL}
\author{ 
    Yihan Wang\textsuperscript{1,2}\thanks{$\ $ Authors contributed equally.} \quad
    Peiyu Liu\textsuperscript{3*} \quad
    Xin Yang\textsuperscript{1}\thanks{$\ $ Corresponding author.} \quad\\
    \textsuperscript{1}China Academy of Information and Communications Technology \\
    \textsuperscript{2}Renmin University of China\quad
    \textsuperscript{3}University of International Business and Economics \\
    \texttt{yihan3123@gmail.com} \quad
    \texttt{liupeiyustu@163.com} \quad
    \texttt{yangxincps@163.com}
}
\usepackage{hyperref}
\begin{document}
\maketitle
\begin{abstract}

Schema linking is a critical bottleneck in applying existing Text-to-SQL models to real-world, large-scale, multi-database environments. Through error analysis, we identify two major challenges in schema linking: (1) \emph{Database Retrieval}: accurately selecting the target database from a large schema pool, while effectively filtering out irrelevant ones; and (2) \emph{Schema Item Grounding}: precisely identifying the relevant tables and columns within complex and often redundant schemas for SQL generation.
Based on these, we introduce LinkAlign, a novel framework tailored for large-scale databases with thousands of fields. LinkAlign comprises three key steps: multi-round semantic enhanced retrieval and irrelevant information isolation for Challenge 1, and schema extraction enhancement for Challenge 2. Each stage supports both Agent and Pipeline execution modes, enabling balancing efficiency and performance via modular design. 
To enable more realistic evaluation, we construct AmbiDB, a synthetic dataset designed to reflect the ambiguity of real-world schema linking. Experiments on widely-used Text-to-SQL benchmarks demonstrate that LinkAlign consistently outperforms existing baselines on all schema linking metrics. Notably, it improves the overall Text-to-SQL pipeline and achieves a new state-of-the-art score of 33.09\% on the Spider 2.0-Lite benchmark using only open-source LLMs, ranking first on the leaderboard at the time of submission.
The codes are available at~\href{https://github.com/Satissss/LinkAlign}{https://github.com/Satissss/LinkAlign}.
\end{abstract}

\section{Introduction}\label{sec:introduction}
\begin{table}[ht]
\centering
\small
\begin{tabular}{p{3.5cm} c c}
\toprule
\textbf{Approach} & \textbf{Open-source} & \textbf{Score (\%)} \\
\midrule
ReFoRCE + o1-preview & \xmark & 30.35 \\
Spider-Agent + Claude-3.7 & \xmark & 25.41 \\
Spider-Agent + o3-mini & \xmark & 23.40 \\
DailSQL + GPT-4o & \xmark & 5.68 \\
CHESS + GPT-4o & \xmark & 3.84 \\
DIN-SQL + GPT-4o & \xmark & 1.46 \\
\midrule
LinkAlign + DeepSeek-R1 & \cmark & \textbf{33.09} \\
LinkAlign + DeepSeek-V3 & \cmark & 24.86 \\
\bottomrule
\end{tabular}
\caption{Comparision across methods on Spider 2.0-lite benchmark. Our method achieves new SOTA score of 33.09 purely using open-source LLMs.}
\label{tab:spider2}
\end{table}
Text-to-SQL~\cite{Zhong2017, Wang2017, Cai2017, Qin2022} aims to enable non-expert users to retrieve data effortlessly by automatically translating natural language questions into accurate SQL queries. 
Recent advances in large language models (LLMs) have led to notable improvements in Text-to-SQL benchmarks~\cite{Sun2023,Pourreza2024_2}, showcasing their growing capabilities in understanding and generating SQL queries. However, existing methods often fall short in real-world enterprise applications due to difficulties in handling massive redundant schemas and complex multi-database environments (\eg local and cloud systems). It faces significant failures in adapting existing Text-to-SQL models to large-scale multi-database scenarios largely due to~\emph{schema linking}, \ie identifying 
the necessary database schemas~(tables and columns) from large volumes of database schemas for user queries~\cite{wang2019rat,guo2019towards,Talaei2024}. The underlying reasons for these failures remain unexplored, leaving a gap in addressing the real-world Text-to-SQL tasks.

To understand the failures, we conduct a systematic error analysis and identify two major challenges underlying the schema linking. 
\emph{Challenge~1 - Database Retrieval} : how to accurately select the target database from a large schema pool, while effectively filtering out irrelevant ones. Existing researches often overlook this challenge as they often assume that schemas from single-database are small-scale and can be directly fed into models for efficient processing. 
\emph{Challenge~2 - Schema Item Grounding} : how to precisely identify the relevant tables and columns within complex and redundant schemas for SQL generation. The post-retrieval phase must handle a large volume of semantically similar tables and columns, which increases the risk of overlooking critical items necessary for generating accurate SQL queries.

Motivated by these factors, we propose LinkAlign, a novel framework that systematically addresses the challenges of schema linking in real-world environment. To address Challenge 1, our approach focuses first on (1)~\emph{Retrieving Potential Database Schemas} through multi-round semantic alignment by query rewriting. This step infers missing database schemas from retrieval results by leveraging the LLM's reflective capabilities, then rewrites the query to align with the ground truth schema semantically. Then our approach centers on (2)~\emph{isolating irrelevant schema information} to reduce noise by response filtering. This step filters out the database noise from a set of candidates, minimizes interference from irrelevant schemas, and streamlines the downstream processing pipeline. To address Challenge 2, our approach directs efforts towards (3)~\emph{extracting schemas for SQL generation} through identifying tables and columns by schema parsing. This step scales schema linking to large-scale databases by introducing advanced reasoning-enhanced prompting techniques like multi-agent debate~\cite{Chan2023,pei2025socratic} and chain-of-thought~\cite{Wei2022}. 
To balance efficiency and accuracy, we propose two complementary implementation paradigms:~\emph{Pipeline} and~\emph{Agent}.
The pipeline mode executes each step via a process-fixed single LLM call, offering a streamlined, low-latency solution ideal for real-time database queries. In contrast, the agent mode performs multi-turn agent collaboration during inference, harnessing test-time computation to scale schema linking capabilities to databases with massive and complex schemas.

To better evaluate the model's schema linking capabilities, we automatically construct AmbiDB, a variant of the Spider~\cite{Yu2018} benchmark, which introduces a large number of complex synonymous databases to simulate the challenges in large-scale, multi-database scenarios. We perform comprehensive evaluations on the Spider, Bird~\cite{Li2024_1} and Spider 2.0~\cite{lei2024spider} benchmarks. The framework consistently outperforms baselines in all schema linking metrics. By applying LinkAlign to the classic DIN-SQL~\cite{Pourreza2024_1} method, the framework achieves a state-of-the-art score of 33.09 on the Spider 2.0-Lite benchmark using only open-source LLMs, highlighting its effectiveness in tackling schema linking challenges and enhancing the performance of the Text-to-SQL pipeline.

\section{Error Analysis}
To better understand the gap between existing research and real-world environments, we evaluate 500 samples from the Spider dataset and analysis common error types when models handling schemas across all databases, rather than limiting the scope to small-scale schemas from a single database. To avoid context-length overflow, we employ vectorized retrieval to extract semantically relevant schemas. The results indicate that schema linking errors are the main cause of Text-to-SQL failures, with an error rate greater than 60\%. We manually examined the erroneous samples and identified four error types, which further highlight the two key challenges presented in the Introduction. More details are provided in Appendix~\ref{app:error_analysis}.



\paratitle{Error 1: Target Database without Retrieval (Database Retrieval).}
This error indicates that the retrieved results do not include complete ground-truth database schemas, accounting for 23.6\% of the failures. 
For example, the user intends to query "which semester the master and the bachelor both got enrolled in", but the retrieved results exclude $degree\_program$ table from the target database, which requires inference based on query semantics and commonsense knowledge. However, general vectorized retrieval approaches only return semantically similar results based on embedding distance, conflicting with the fact that user complex queries often misaligned with the ground-truth schema.

\paratitle{Error 2: Referring Irrelevant Databases (Database Retrieval).}
Unlike Error 1 that focuses on the missing target database, Error 2 centers on the irrelevant schema noise introduced by imprecise retrieval.
This error indicates that the model refers to incorrect schemas from unrelated databases when generating SQL, accounting for 13.3\% of the failures.
For example, the user intends to query "The first name of students who have both cats and dogs". However,the generated SQL incorrectly infers $People.first\_name$ from an unrelated database while overlooking the ground truth $Student.f\_name$, even though both are successfully retrieved and fed into the model. 
Without isolating irrelevant database information, the model tends to mistakenly select seemingly more appropriate schemas from unrelated databases, leading to SQL execution failures.

In summary, Error 1 and Error 2 highlight the gap between existing methods and real-world large-scale multi-database scenarios. Even though these are avoided, SQL execution still fails due to other schema linking errors.

\paratitle{Error 3: Linking to the Wrong Tables (Schema Item Grounding).}
This error indicates that the generated SQL omits or misuses tables, accounting for 19.8\% of the failures. 
For example, both the $student$ and $people$ tables have fields for $name$, but 
the model selects the latter incorrectly.
In real-world scenarios, the models often
overlook critical tables, which directly impairs the execution accuracy of the generated SQL~\cite{Maamari2024}. 

\paratitle{Error 4: Linking to the Wrong Columns (Schema Item Grounding).}
This error indicates that the generated SQL omits or misuses fields despite referencing the ground truth table correctly, accounting for 11.6\% of the failures. For example, the model may omit the join columns $pets.pet\_id$ and $has\_pet.pet\_id$ in the join operation of the correct SQL statement. Missing such critical columns directly incurs SQL execution failure.


\begin{figure*}[ht]
    \centering
    \includegraphics[width=\textwidth]{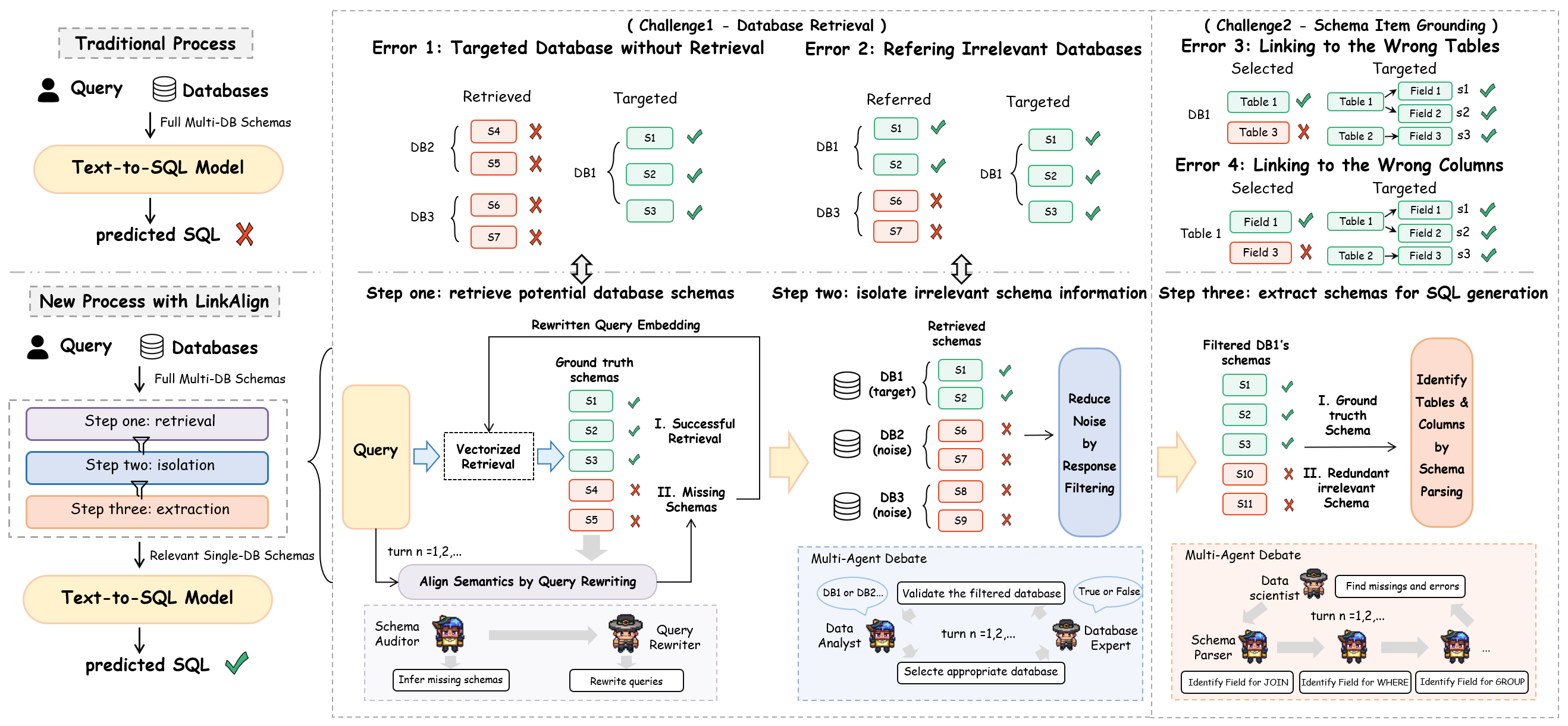}  
    \caption{Overview of the LinkAlign framework including three core components.}
    \label{fig:overview}
\end{figure*}

\section{Methodology}\label{sec:methods}
This section introduces the LinkAlign framework, scaling schema linking to large-scale, multi-database environments through three key steps. The framework begins by (1) retrieving potential database schemas via multi-round semantic alignment through query rewriting, effectively recalling the ground-truth schemas while significantly reducing the candidate pool. Next, (2) isolates irrelevant schema information through response filtering, enabling precise target database localization and noise reduction by discarding unrelated candidates. Finally, it focuses on (3) extracting schemas for SQL generation by identifying necessary tables and columns through schema parsing. To balance efficiency and effectiveness, we provide two complementary implementation modes—Pipeline and Agent—for each step of the framework.

\subsection{Background}
Before proposing our method, we consider a typical Text-to-SQL setting. Given a set of \( N \) databases \( D = \{ D_1, D_2, \dots, D_N \} \) and the schemas \( S = \{ S_1, S_2, \dots, S_N \} \), where a schema \( S_i \) is defined as \( S_i = \{ T_i, C_i \} \), with \( T_i \) representing multiple tables \( \{ T_1^i, T_2^i, \dots, T_{|T_i|}^i \} \) and \( C_i \) representing columns \( \{ C_1^i, C_2^i, \dots, C_{|C_i|}^i \} \). 
Traditional methods take full multi-database schemas and user query as input and rely on schema linking component to identify tables and columns for SQL generation:
\begin{equation}
    S^\prime = f_{\text{parser}} \left( S, Q, c \mid E, LLM \right),
\end{equation}
where \(f_{\text{parser}} \left(\cdot \mid E \right)\) denotes the schema parsing function based on the text-embedding model $E$ and LLM. Symbol $c$ denotes additional context such as field descriptions or sampling examples.

\subsection{Modular Step Design}\label{sec:step_illustration}
This section outlines the modular design of each step, decoupling from implementations to accommodate diverse application scenarios. 

\paratitle{Step one: retrieve potential database schemas.}
To mitigate the exclusion of the ground truth schema (Error 1), we propose a multi-round semantically enhanced retrieval method to recall critical schemas without significantly increasing the retrieval size. This step infers missing schemas from retrieval results by leveraging the LLM's reflective capabilities, then rewrites the query to align with the ground-truth schema semantically. 

Specifically, following each retrieval round, field-level metadata (e.g., type, description, value example) from index nodes are extracted and serialized into structured natural language sequences aligned with LLM processing preferences. The resulting schema representation, combined with the original user query, forms the context, denoted as the tuple ($S_{r_i},\ Q_0$). Subsequently, we leverage LLMs to evaluate the semantic alignment between the user query and the retrieved schema context, and further infer potentially missing schema elements critical for accurate SQL generation. The inferred schemas are integrated with the original query and optimized to remove redundant or ambiguous expressions. This integration helps reduce hallucination-induced deviations from user intent and improves semantic alignment with the ground-truth schema. The rewritten queries are then embedded into vector representations and used to retrieve relevant database schemas. Finally, the retrieval results are ranked and aggregated based on the number of rewrites and their similarity scores:
\begin{equation}
    Z = \bigcup_{t=0}^{T} f_{\text{retriever}} \left( S, Q_t, c\mid E \right),
\end{equation} 
where \(T\) represents the number of query rewrites. By rewriting queries and enhancing their semantic representation, this approach improves the alignment between user queries and database schemas, ensuring more accurate retrieval outcomes. Dynamically adjusting the retrieval strategy based on feedback enables high retrieval performance with fewer iterations. Concurrently, multi-round iterative optimization enables effective scalability to large-scale databases with massive schemas. Here is an example of the query rewriting process.
\begin{newshaded}
\small
\noindent \ding{227} {\bf User Query $Q_0$}: Which semester the master and the bachelor both got enrolled in?

\noindent {\textbf{Missing Schema}}:~\texttt{degree\_programs (degree\_type)}
\noindent [1] {\bf Rewrite $Q_1$}: In a database with \texttt{degree\_programs}, how to find semesters where both master's \texttt{degree\_type} and bachelor's \texttt{degree\_type} programs exist? Group by \texttt{enrollment\_semester} with checks for both program types.
\noindent \rule{\textwidth}{0.2pt} 
\noindent {\textbf{Missing Schema}}:~\texttt{enrollment\_records (semester)}

\noindent [2] {\bf Rewrite $Q_2$}: In a database with \texttt{enrollment\_records}, how to find semesters where both master and bachelor students enrolled? Group by \texttt{semester} and filter for overlapping enrollments.
\end{newshaded}


\paratitle{Step two: isolate irrelevant schema information.}
While multi-round retrieval substantially enhances the recall of critical schema elements, embedding-based similarity comparisons are prone to introducing additional semantically proximate but irrelevant noise (Error 2). To mitigate this challenge, we introduce a filtering mechanism designed to prune redundant or irrelevant schema elements. Although we prioritize target database localization in multi-database settings, which challenges non-technical users who lack expert database knowledge, the framework remains effective in filtering out noise in single-database settings, which can serve as a subsequent operation after localization. To further improve performance in single-database settings, we propose two optimization strategies: Random Preservation with Exponential Decay and Post-Retrieval, detailed in the Appendix~\ref{sec:single_db}. We now focus on the multi-database setting.

Once the retrieved results $Z$ contain schemas from irrelevant databases, the next step is to precisely locate the target database $D_t$ while filtering out irrelevant ones. To achieve this, the framework initially groups all schemas by their respective databases, enabling subsequent processing to treat each database as a cohesive unit. The framework then compares the relevance of each candidate database $D_i$ by evaluating how well its associated schemas satisfy the user's query intent and then ranks these databases accordingly. The database exhibiting the highest relevance, $D_t$, is then designated as the target database, concurrently with the suppression of schema noise originating from irrelevant databases.
\begin{equation}
    D_{t} = \arg\max_{1<i<N} P_M(D_i \mid Q_0, Z),
\end{equation}
where M denotes the LLMs used for analysis.
By isolates unrelated schemas information, this step enables subsequent processes to concentrate computational resources on the most appropriate database, improving schema linking performance while achieving cost efficiency.

\paratitle{Step three: extract schemas for SQL generation.}
To mitigate schema misuse during SQL generation (Error 3 and Error 4), it is imperative to precisely parse the schema and identify the most relevant tables and columns. This procedure emulates the laborious manual process of schema extraction, yet it is fully automated by leveraging the intrinsic knowledge and reasoning capabilities of LLMs. Crucially, this approach scales schema linking to complex, redundant, and large-scale databases through the integration of advanced reasoning-enhanced prompting techniques, including multi-agent debate and chain-of-thought reasoning.

Specifically, from the filtered database schema $S_{\hat{u}}$ derived in the preceding steps, the objective is to identify a salient subset $S_{\hat{u}}^\prime$ comprising the most relevant tables $T_i^{\hat{u}}$  and columns $C_i^{\hat{u}}$. This selection is predicated on their alignment with the user query, thereby ensuring the resulting schema is both precise and comprehensively representative.
\begin{equation}
    S_{\hat{u}}^\prime = \{ T_i^{\widehat{u}},C_i^{\widehat{u}} |\ \mathbb{I}(Q_0,C_i^{\widehat{u}}) =1 \}\},
\end{equation}
where\(\mathbb{I} (\cdot) \)is an abstract indicator that determines whether a column is needed based on the query, returning 1 if true and 0 otherwise. In stark contrast to traditional text-to-SQL approaches, which typically depend on static mappings, the LinkAlign framework leverages dynamic reasoning to robustly address intricate schema-linking challenges.

\subsection{Component Implementation Optimization}
Drawing upon the modular definitions outlined in Section~\ref{sec:step_illustration}, we introduce two distinct strategies to implement the core components of each step, as depicted by the dashed boxes in Figure~\ref{fig:overview}. This modular framework enables flexible combinations based on specific query scenarios, allowing for optimized trade-offs between computational cost and processing effectiveness. 

The first strategy is the Single-Prompt Pipeline, which executes each step through a single process-fixed LLM call. This design offers a low-latency streamlined solution, making it ideal for real-time database queries. A detailed exposition of this strategy is provided in Appendix~\ref{method_explanation}. Conversely, this section will primarily focus on the Multi-agent Collaboration strategy. This approach prioritizes accuracy and offers robust capabilities for tackling complex query tasks in real-world environments.

\paratitle{Align Semantics by Query Rewriting.} 
Inspired by the reflective capabilities of LLMs demonstrated by \citeauthor{Shinn2024}, we introduce a semantic-enhanced retrieval approach based on retrieval feedback to achieve precise alignment with the ground-truth schema. Specifically, the Schema Auditor initiates by mapping the user queries into structured triplets (entities, attributes, and constraints). Next, it scrutinizes the retrieval results to infer missing schemas that may critical for accurate SQL generation (e.g., tables or fields for SELECT, JOIN, or WHERE clauses). This process culminates in an audit report that summarizes the parsed query, the inferred missing schemas, and the corresponding confidence levels.  Subsequently, the Query Rewriter Agent leverages the comprehensive report to enhance the original query by clarifying ambiguous expressions, supplementing semantic information for missing elements, and transforming the query into a template format optimized for text embedding models, ultimately improving retrieval recall for the ground-truth schema.

\paratitle{Reduce Noise by Response Filtering.} 
When multiple candidate schemas exhibit minimal semantic differentiation, achieving consensus through multi-agent debate can significantly mitigate the risk of confusion.
Inspired by this insight, we meticulously designed a multi-agent debate model comprising two distinct LLM agents: Data Analyst and Database Expert. Specifically, the Data Analyst evaluates the alignment between each database and the user query based on domain relevance and schema coverage completeness, then ranking them through corresponding comprehensive assessment. The highest ranked database is then selected from all candidates, with its schema and query context provided to the Database Expert. Subsequently, the Database Expert rigorously evaluates whether its provided database schema can satisfy the query requirements, validating the selection's appropriateness and determining whether to retain it. The debate follows a one-by-one strategy, \ie starting with the data analyst, after which the two roles present their perspectives in turn. The debate ends when a predefined number of rounds is reached, and then a terminator outputs the consensus database as a final result.

\paratitle{Identify Tables and Columns by Schema Parsing.} To enhance schema linking capabilities in complex scenarios, we meticulously designed a Multi-Agent Debate framework comprising two distinct LLM agents: Schema Parser and Data Scientist. Specifically, the Schema Parser extracts potentially required schema elements across three dimensions — tables, fields, and relationships — conducting reviews to prevent omissions. Extraction results from multiple Schema Parsers are then aggregated and submitted to the Data Scientist, who subsequently verifies all results, identifying any omissions or errors. The debate process follows a Simultaneous-Talk-with-Summarizer strategy, wherein multiple peer Schema Parsers engage in concurrent deliberation during each round, with final outcomes evaluated by the authoritative Data Scientist. Multi-role participation enhances the recall of tables and columns required for SQL generation, with diverse answers complementing each other to reduce the randomness of single-prompt outputs. 

\begin{table*}[!h]
\centering
\resizebox{11.5cm}{!}{
\begin{tabular}{p{2.5cm}ccccccccc}
\toprule
\textbf{Approach} \centering & \multicolumn{3}{c}{\textbf{Spider}} & \multicolumn{3}{c}{\textbf{Bird}} & \multicolumn{3}{c}{\textbf{AmbiDB}} \\
\cmidrule(lr){2-4} \cmidrule(lr){5-7} \cmidrule(lr){8-10}
 & LA & EM & Recall & LA & EM & Recall & LA & EM & Recall \\
\midrule
LlamaIndex \\
DIN-SQL\centering & 80.0 & 26.8 & 62.4 & 68.8 & 5.1 & 31.3 & 59.7 & 13.3 & 44.2 \\
PET-SQL\centering & 84.1 & 38.6 & 67.2 & 77.1 & 8.2 & 39.7 & 66.4 & 22.0 & 50.2 \\
MAC-SQL\centering  & 82.3 & 17.3 & 42.8 & 75.0 & 5.7 & 34.5 & 65.1 & 9.7 & 30.8 \\
MCS-SQL \centering& 81.0 & 24.3 & 73.2 & 73.7 & 13.9 & 56.1 & 61.9 & 13.7 & 54.8 \\
RSL-SQL \centering & 74.8 & 29.1 & 76.1 & 80.0 & 16.1 & 61.8 & 62.4 & 17.9 & \textbf{59.6} \\
\midrule
Pipeline(ours)\centering & 85.4 & 37.4 & 65.9 & 66.8 & 8.6 & 38.1 & 69.4 & 20.3 & 50.4 \\
Agent(ours)\centering & \textbf{86.4} & \textbf{47.7} & \textbf{80.7} & \textbf{83.4} & \textbf{22.1} & \textbf{64.9} & \textbf{67.6} & \textbf{22.4} & 56.9 \\
\bottomrule
\end{tabular}
}
\caption{Comparison of LA, EM and Recall across different methods in multi-database scenario.}
\label{multi_results}
\end{table*}

\begin{table*}[!h]
\centering
\renewcommand{\arraystretch}{1}
\resizebox{13cm}{!}{
\begin{tabular}{p{2.5cm}ccccccccc}
\toprule
\textbf{Approach} \centering & \multicolumn{3}{c}{\textbf{Spider-dev}} & \multicolumn{3}{c}{\textbf{Bird-dev}} & \multicolumn{3}{c}{\textbf{AmbiDB}} \\
\cmidrule(lr){2-4} \cmidrule(lr){5-7} \cmidrule(lr){8-10}
 & Precision & Recall & EM & Precision & Recall & EM & Precision & Recall & EM \\
\midrule
DIN-SQL \centering& 83.9 & 73.2 & 40.4 & 79.9 & 55.7 & 13.1 & 86.6 & 76.9 & 44.2 \\
PET-SQL \centering & \textbf{84.8} & 73.9 & 33.4 & \textbf{81.6} & 64.9 & 25.9 & \textbf{90.2} & 78.3 & 39.5 \\
MAC-SQL  \centering& 75.0 & 66.8 & 24.4 & 76.3 & 56.2 & 9.1 & 79.8 & 69.6 & 30.1 \\
MCS-SQL \centering & 66.7 & 85.0 & 29.8 & 79.6 & 76.9 & 25.5 & 71.5 & \textbf{88.5} & 34.1 \\
RSL-SQL  \centering & 74.8 & 84.3 & 37.6 & 78.1 & 77.5 & 27.7 & 80.7 & 88.3 & 42.2 \\
\midrule
Agent (ours) \centering & 80.2 & \textbf{87.3} & \textbf{48.1} & 77.1 & \textbf{79.4} & \textbf{29.0} & 86.7 & 85.8 & \textbf{51.5} \\
\bottomrule
\end{tabular}
}
\caption{Comparison of Precision, Recall and EM across different methods in single-database scenario.}
\label{tab:single_results}
\end{table*}

\section{Experiments}\label{sec:experiments}
\subsection{Experimental Setup}
\paratitle{Dataset.} We evaluate our method performance of schema linking on the SPIDER, BIRD and AmbiDB datasets, and the ability to adapt existing Text-to-SQL models to real-world environments on the SPIDER 2.0 benchmark. We provide more details about SPIDER, BIRD and SPIDER 2.0 in Appendix ~\ref{comparable_benchmark}, and the construction of the AmbiDB dataset in Appendix ~\ref{AmbiDB_construction}.

\paratitle{Baselines.} We compare our method against multiple LLM-based schema linking methods. DIN-SQL~\cite{Pourreza2024_1} employs a prompt-driven approach with a single LLM call and a chain-of-thought strategy to improve reasoning. PET-SQL~\cite{Li2024_2} generates preliminary SQL to infer schema references. MAC-SQL~\cite{Wang2024} uses a Selector agent to identify minimal relevant schema subsets. MCS-SQL~\cite{Lee2024} applies a two-step table and column linking process with multiple prompts and random shuffling for robustness. RSL-SQL~\cite{cao2024rsl} adopts a bidirectional linking strategy that combines forward and backward schema linking.





\paratitle{Metrics.}\label{evaluation_metric} We evaluate the ability of schema linking using the following metrics:

$\bullet$~\textbf{Locate Accuracy~(LA).} Let $N$ denote the total number of test examples and $N_a$ the number of examples without Error 1 or Error 2. The LA is defined as $N_a$ / $N$, measuring the model's ability to locate the target database accurately.

$\bullet$~\textbf{Exact Matching~(EM).} Let $N_e$ denote the number of examples without Error 1 to 4. The EM score is defined as $N_e$ / $N$, measuring the model's ability to perform precise schema linking. 

$\bullet$~\textbf{Recall.} This metric measures the recall of database schemas from the ground truth SQL. It is preferred over Precision, as minor schema noise may not significantly impact SQL generation~\cite{Maamari2024}. But it still makes sense for the model to maintain a high recall rate while improving precision, in order to minimizing noise introduced by excessive irrelevant schema.

We evaluate the ability of Text-to-SQL using the Execution Accuracy:

$\bullet$~\textbf{Execution Accuracy(EX).} This metric is widely used to evaluate the quality of the generated SQL~\cite{Yu2018,Li2024_1,lei2024spider}, based on the comparison with the results of the Gold SQL execution.

\paratitle{Implementations.} The open-sourced text-embedding model bge-large-en-v1.5 is applied to convert database schema metadata and queries into vectors. We set the top\_k of the retrieval size at 5 and adaptively adjust turn\_n according to the database size. We use \texttt{GLM-4-air} model for schema linking and \texttt{DeepSeek-V3, R1 and Qwen-72B} for "end-to-end" Text-to-SQL evaluation. We further developed a versatile Text-to-SQL development and evaluation tool, enabling multi-task concurrent calls via configuration files, thereby supporting subsequent experimental testing.

\subsection{Main Results}
\subsubsection{Schema Linking Performance}
\paratitle{Multi-Databases Results.} As shown in Table ~\ref{multi_results}, our method achieves the highest Locate Accuracy (LA) on the SPIDER, BIRD, and AmbiDB datasets, with scores of 86.4\%, 83.4\%, and 69.4\%, respectively, demonstrating the effectiveness in mapping data requirements to the target database. A key contributor to this improvement is the introduction of the Response Filtering step, which mitigates Error 2 by eliminating irrelevant database schemas. Additionally, our framework achieves the highest Exact Match (EM) performance across all three datasets, outperforming baseline models by margins of 23.6\%, 1.8\%, and 37.3\%, respectively.The presence of error 1,2 in multi-database contexts makes it difficult to avoid blending unrelated schemas, leading to challenges in accurately recalling relevant tables and columns. 
Especially as the database size increased, we observed a significant decrease in the recall rate of all methods on the AmbiDB dataset, further demonstrating that our proposed dataset exacerbated the challenge.

\paratitle{Single-Database results.} We further compare and evaluate the ability of different methods to identify correct tables and columns in a large-scale database. As shown in Table ~\ref{tab:single_results}, our Agent method achieves state-of-the-art performance across all datasets in terms of the EM evaluation metric. After eliminating interference from irrelevant database schemas, the recall rate for all methods significantly improved compared to the results in Table ~\ref{multi_results}. Compared to the baseline models, the Agent method achieves the highest recall rates on both the Spider-dev and Bird-dev datasets. This demonstrates that our method exhibits superior performance and robustness when the inference capabilities of large models are limited. Although, on the AmbiDB dataset, our method's recall rate lags behind MCS-SQL (88.5\%) and RSL-SQL (88.3\%), our method outperforms these models by 21.3\% and 7.4\%, respectively, in Precision. This indicates that our method maintains high recall while minimizing irrelevant noise. Overall, considering all metrics, our method demonstrates excellent performance and robust schema-linking capabilities.

\subsubsection{Text-to-SQL Performance}
\paratitle{Spider 2.0-lite Results.}
To convincingly demonstrate the effectiveness of the framework, we conducted tests on the Spider 2.0-Lite benchmark~\cite{lei2024spider}, which simulates real-world challenges by significantly increasing the number of schemas. As shown in Table~\ref{tab:spider2}, we achieved the new SOTA score of 33.09\% applying LinkAlign to the DIN-SQL method of lowest rank. In particular, our method achieves performance comparable to the existing baseline like ReFoRCE and Spider-Agent using purely open-source LLMs. The results highlight the framework's effectiveness in enhancing Text-to-SQL performance by improving schema linking in large-scale database environments.


\begin{table}[!h]
\centering
\small
\begin{tabular}{p{4.5cm} >{\centering\arraybackslash}p{2cm}}
\toprule
\textbf{Approach} & \textbf{EX (\%)} \\
\midrule
DIN-SQL + GPT-4 & 82.8 \\
MAC-SQL + GPT-4 & 86.8 \\
DAIL-SQL + GPT-4 & 86.6 \\
MCS-SQL + GPT-4 & 89.5 \\
\midrule
LinkAlign$^*$ + GPT-4 & 91.2 \\
LinkAlign$^*$ + DeepSeek-V3(671B) & 88.9 \\
LinkAlign$^*$ + Qwen(72B) & 86.8 \\
\bottomrule
\end{tabular}
\caption{Comparison of different methods on Spider-dev dataset. $^*$ indicates method using a simplified LinkAlign framework without Step One and Step Two.}
\label{tab:spider}
\end{table}

\begin{table}[!h]
\centering
\small
\begin{tabular}{p{4.5cm} >{\centering\arraybackslash}p{2cm}}
\toprule
\textbf{Approach} & \textbf{EX (\%)} \\
\midrule
DIN-SQL + GPT-4 & 50.7 \\
MAC-SQL + GPT-4 & 59.4 \\
DAIL-SQL + GPT-4 & 54.8 \\
MCS-SQL + GPT-4 & 63.4 \\
RSL-SQL + GPT-4 & 67.2 \\
\midrule
LinkAlign$^*$ + GPT-4 & 61.6 \\
LinkAlign$^*$ + DeepSeek-V3(671B) & 57.5 \\
LinkAlign$^*$ + Qwen(72B) & 53.4 \\
\bottomrule
\end{tabular}
\caption{Comparison of different methods on Bird-dev dataset. $^*$ indicates method using a simplified LinkAlign framework without Step One and Step Two.}
\label{tab:bird}
\end{table}
\paratitle{Small-scale database results.}
We assess the framework ability to generalize improved schema linking to smaller-scale databases by evaluating it on the Spider and Bird dev set. To further assess generalization across diverse LLMs, we employed two open-source models, Deepseek and Qwen, with significantly different parameter sizes of 671B and 72B, respectively. Given that the limited database schemas would not exceed the model's context and minor redundant schema noise would not impact the LLMs' attention significantly, we only utilized a
simplified LinkAlign framework by excluding Step One and Step Two. The results show that Execution Accuracy gains of 6.7\% on Spider and 7.2\% on Bird, demonstrating that improved schema linking enhances SQL generation significantly.

\subsection{Runtime Efficiency}
We assessed the average runtime of each step in LinkAlign using samples from the Spider 2.0-lite dataset. 
The results show that pipeline mode is more efficient and better suited for latency-sensitive scenarios, while agent mode offers improved performance when accuracy is prioritized. This flexibility enables users to adapt LinkAlign to different application needs.

\begin{table}[ht]
\centering
\small 
\begin{tabular}{
    >{\raggedright\arraybackslash}p{1.4cm}  
    >{\centering\arraybackslash}p{1.5cm}    
    >{\centering\arraybackslash}p{1.5cm}    
    >{\centering\arraybackslash}p{1.5cm}    
}
\toprule
\textbf{Approaches} & \textbf{S1 Time (s)} & \textbf{S2 Time (s)} & \textbf{S3 Time (s)} \\
\midrule
Pipeline & 9.02 & 2.94 & 1.67 \\
Agent    & 30.90 & 26.23 & 13.46 \\
\bottomrule
\end{tabular}
\caption{Average time for each step of the framework.}
\label{tab:efficiency}
\end{table}

\begin{table}[!h]
\centering
\resizebox{\columnwidth}{!}{
\small
\begin{tabular}{lcccccc}
\toprule
{\textbf{Model variant }} \centering  & \multicolumn{3}{c}  {\textbf{Spider}} & \multicolumn{3}{c}{\textbf{AmbiDB}} \\
\cmidrule(lr){2-4}\cmidrule(lr){5-7}
 & \textbf{LA} & \textbf{EM} & \textbf{Recall} & \textbf{LA} & \textbf{EM} & \textbf{Recall} \\
\midrule
\textbf{Pipeline} \centering & 85.4 & 37.5 & 66.1 & 69.4 & 20.3 & 50.4 \\
\textbf{w/o que. rew.} \centering & 85.3 & 37.7 & 72.3 & 63.1 & 14.5 & 52.8 \\
\textbf{w/o res. fil.} \centering & 81.9 & 26.0 & 62.0 & 66.2 & 15.3 & 48.5 \\
\textbf{w/o both} \centering & 80.0 & 26.8 & 62.4 & 59.5 & 11.4 & 38.7 \\
\midrule
\textbf{Agent} \centering & 86.4 & 47.7 & 80.7 & 67.6 & 22.4 & 56.9 \\
\textbf{w/o que. rew.} \centering & 83.6 & 30.6 & 73.0 & 65.3 & 15.1 & 57.0 \\
\textbf{w/o res. fil.} \centering & 66.7 & 27.8 & 54.8 & 58.5 & 14.5 & 60.6 \\
\textbf{w/o both} \centering & 73.6 & 32.9 & 61.1 & 58.0 & 17.6 & 47.8 \\
\bottomrule
\end{tabular}
}
\caption{Performance comparison of model variants on Spider and AmbiDB datasets. ``que. rew.'' indicates query rewriting and ``res. fil.'' denotes the response filtering.}
\label{tab:model_variants}
\end{table}

\begin{figure*}[!h]  
    \centering
    \begin{minipage}{0.48\linewidth}
        \centering
        \includegraphics[width=\linewidth]{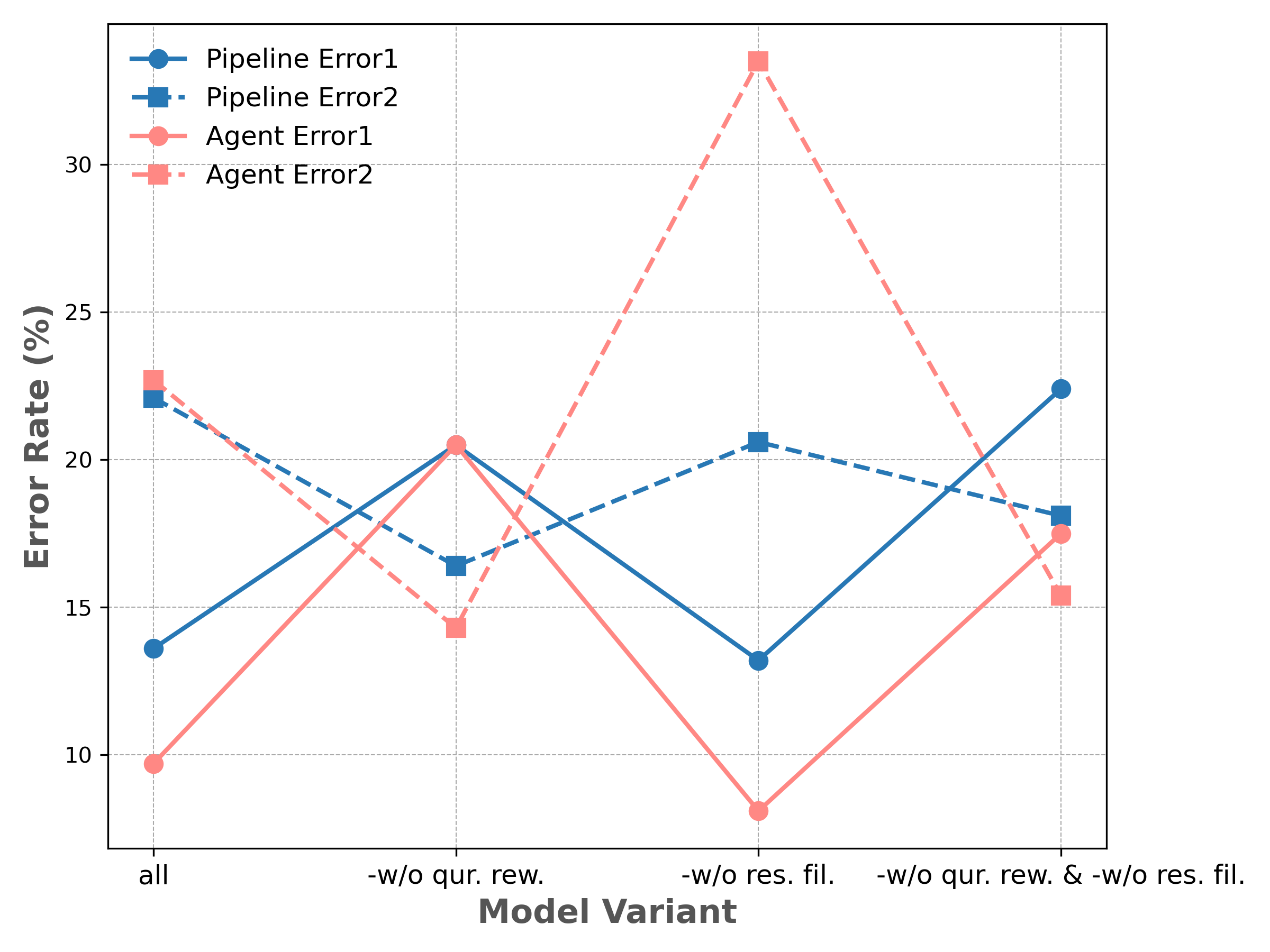}
        \caption{The impact on Error Rates.}
        \label{fig:ablation_error}
    \end{minipage}
    \hfill
    \begin{minipage}{0.48\linewidth}
        \centering
        \includegraphics[width=\linewidth]{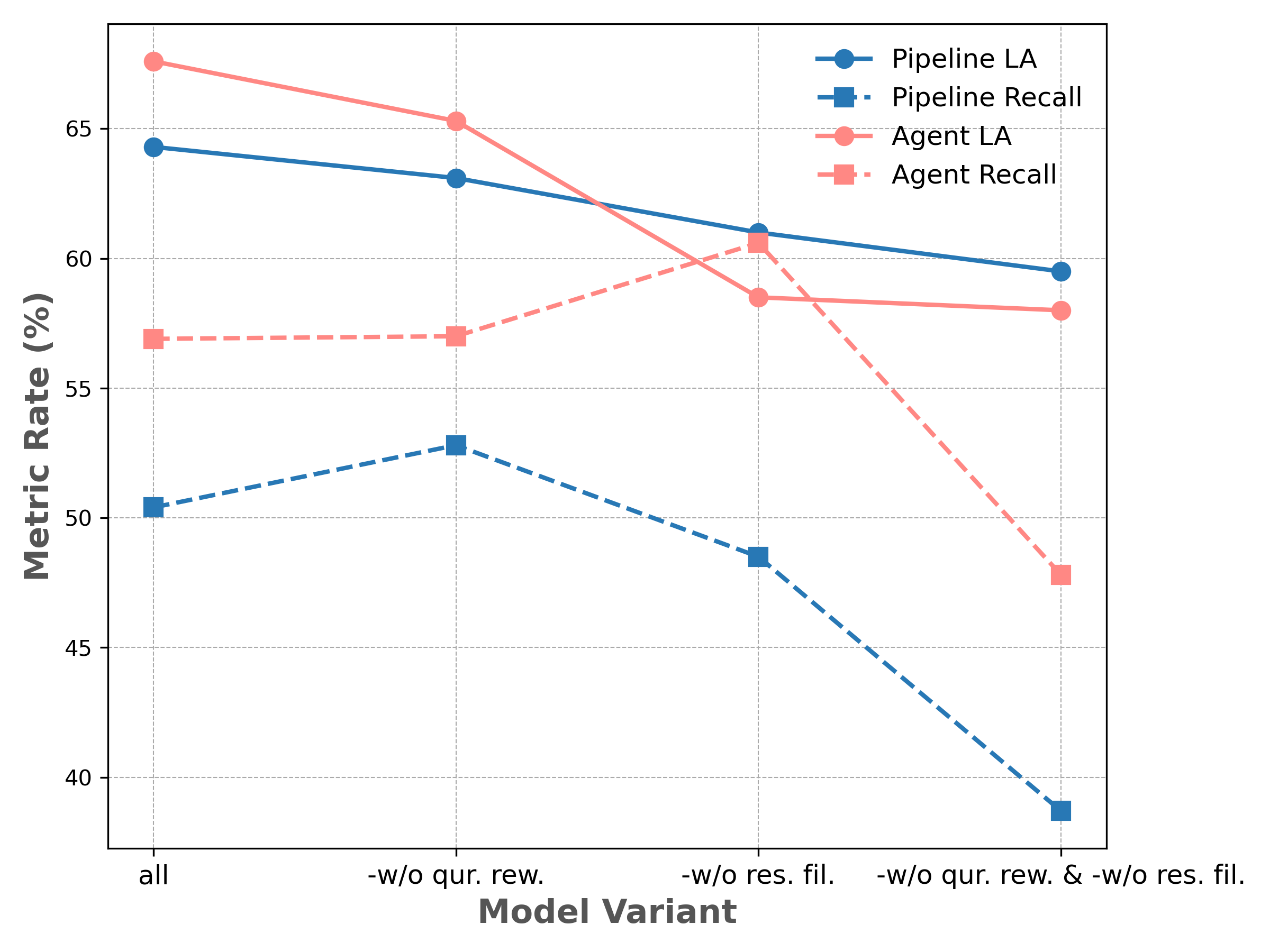}
        \caption{The impact on Evaluation Metrics.}
        \label{fig:ablation_metric}
    \end{minipage}
    \label{fig:overall}
\end{figure*}

\subsection{Ablation Study}
We conducted an ablation study to examine the incremental impact of the two core steps in the LinkAlign framework. We exclude Step 3 from consideration, as schema parsing is often considered fundamental to schema linking and must be retained. As shown in Table~\ref{tab:model_variants}, each step contributes to achieving SOTA performance on the benchmark.

\paratitle{Impact of Query Rewriting.} User queries often misaligned with the target schema, leading to retrieval inefficiency. As shown in Figure~\ref{fig:ablation_error}, adding Query Rewriting reduces Error 1 by 6.9\% and 10.8\% for two mode, improving recall by resolving ambiguity. The Agent mode benefits more than Pipeline, indicating LLM reflection better aligns queries to schemas. However, this step also introduces irrelevant schemas, increasing Error 2 by 5.7\% and 8.4\%, which complicates database localization. Despite this, the net effect is positive: Locate Accuracy improves overall. The improvement is more notable on the AmbiDB dataset than on Spider, showing query rewriting is especially important with higher ambiguity. For simple queries, balancing gains and drawbacks of rewriting is recommended to optimize Locate Accuracy.

\paratitle{Impact of Response Filtering.} Figure~\ref{fig:ablation_error} shows that although Query Rewriting introduces irrelevant databases, Response Filtering reduces Error 2 in the Agent mode by 10.8\%, effectively offsetting this negative effect. The Agent mode gains more than Pipeline, highlighting the filtering step’s critical role in narrowing LLM focus to the correct schema. As Table~\ref{tab:model_variants} demonstrates, Response Filtering improves both EM and Recall for both strategies. Despite wrong database selection causing Schema Linking failures, filtering significantly boosts schema linking by mitigating Error 3 and 4.

\section{Conclusion}
In this paper, we aim to adapt existing methods to real-world large-scale multi-database scenario by tackling the challenge of schema linking. First, we highlight four core errors leading to schema linking failures. Based on this analysis, we propose the LinkAlign framework, which composes of three key steps. Additionally, we introduce the AmbiDB dataset, for better design and evaluation of the schema linking component. Experiments demonstrate that our model outperforms existing baseline methods across all evaluation metrics in both multi-database and single-database contexts.

\section*{Acknowledgments}
This work was partially supported by National Natural Science Foundation of China under Grant No. 62506077.

\bibliography{custom}

\appendix
\section{Related Work} 
\paratitle{Early Work in Text-to-SQL tasks.} Early approaches in Text-to-SQL systems primarily rely on rule-based methods~\cite{zelle1996,saha2016athena}, where manually predefined templates are designed to capture relationships between user queries and schema elements. The development of neural network-based approaches, particularly seq2seq architectures such as LSTM~\cite{hochreiter1997lstm,sutskever2014sequence} and transformers~\cite{vaswani2017attention} significantly improve text-to-sql performance~\cite{choi2021ryansql}. Models like IRNet~\cite{guo2019towards} and RASAT~\cite{qi2022rasat} leverage attention mechanisms to better incorporate schema elements into the query understanding process. A notable breakthrough in schema linking is achieved with the introduction of graph neural networks~\cite{wang2019rat,cao2021lgesql,bogin2019global}, enabling models to represent relational database structure as graphs and learn connections between query and schema elements more effectively. 

\paratitle{Evolution Techniques elicited by LMs.} With the development of pre-trained language models such as BERT~\cite{devlin2018bert} and T5~\cite{raffel2020t5} , approaches such as SADGA~\cite{cai2021sadga} and PICARD~\cite{scholak2021picard} combine PLMs with task-specific fine-tuning to enhance execution accuracy. Recent studies also explore the integration of LLMs~\cite{rajkumar2022,liu2023chatgpt,guo2023prompting}, such as GPT, which can directly generate SQLs without requiring task-specific training data. Building on LLMs, models like C3~\cite{Dong2023}, DIN-SQL~\cite{Pourreza2024_1}, and MAC-SQL~\cite{Wang2024} leverage task decomposition strategies and advanced reasoning techniques, such as Chain-of-Thought~\cite{Wei2022}, Least to Most~\cite{Zhou2022_1} and self-consistency decoding~\cite{wang2022self_consistency} to address  schema linking tasks more effectively.

\paratitle{Emerging Solutions and Challenges.} Schema linking remains a challenge when handling large-scale schema and further complicated by the ambiguity in user queries. Approaches such as CHESS~\cite{Talaei2024} and MCS-SQL~\cite{Lee2024} try to address this by using multiple intricate prompts and sampling responses from LLMs on existing benchmarks such as Spider~\cite{Yu2018} and Bird~\cite{Li2024_1}. To address challenges, some works focus on system robustness and mitigating model hallucinations. For instance, Solid-SQL~\cite{liu2024solid} enhances robustness via a specialized pre-processing pipeline, while TA-SQL~\cite{qu2024tasql} introduces a ``Task Alignment'' strategy to reduce hallucinations by reframing sub-tasks into more familiar problems for the model. Although effective, they still require considerable computational resources and API costs, particularly in large-scale database scenarios.
An alternative line of research is to explore traditional model compression techniques—such as pruning~\cite{xia2024sheared,liu2025masks}, quantization~\cite{lin2024awq,liu2024unlocking,liu2024do}, or distillation~\cite{hsieh2023distilling}, to adapt LLMs into smaller~\cite{liu2021enabling}, task-specific variants. This may provide a cost-efficient and scalable direction for future work.

\section{Error Analysis}\label{app:error_analysis}
To figure out why existing methods fail in real-world environments, we tested 500 examples randomly sampled  from the SPIDER dataset. In particular, the model needs to handle schemas from all databases, rather than small-scale schemas from a single database. Given the challenges existing methods face in handling large-scale database schemas, we adopt a vectorized retrieval approach to search relevant schemas based on user queries. Then the retrieve results composed of related schemas from different databases are fed into DIN-SQL models to generate SQL for user queries.

\begin{figure}[ht]
    \centering
    \includegraphics[width=0.5\textwidth]{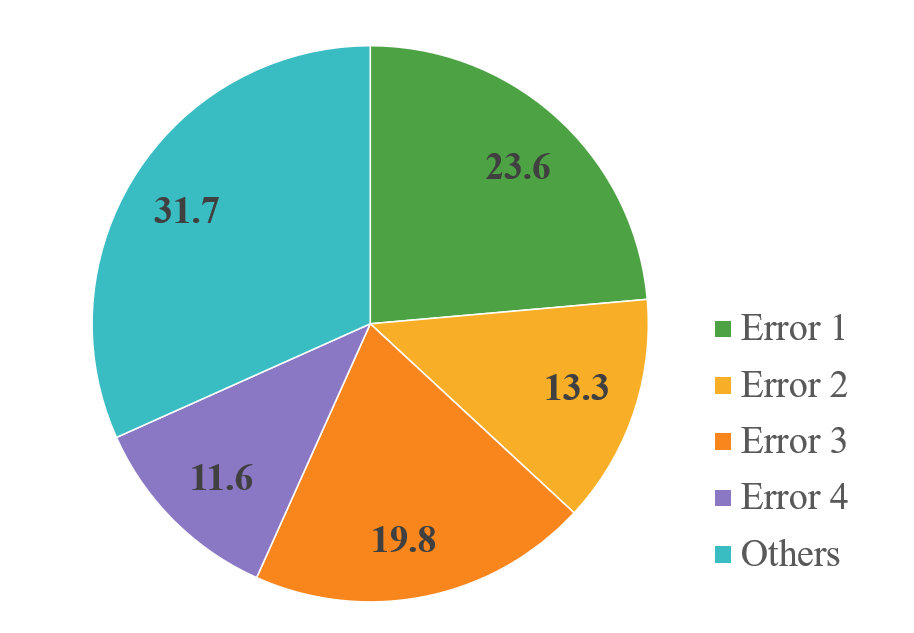}  
    \caption{Error Distribution in Failed Cases.}
    \label{fig:error_analysis}
\end{figure}

Experimental results in Figure~\ref{fig:error_analysis} show that in large-scale, multi-database scenarios, the EX score of the DIN-SQL model drops from 85.3 to 67.4\%. Manual analysis of error cases reveals that schema linking errors account for 68.3\% of Text-to-SQL failures, making them the major failure cause.

\section{Single-Prompt Pipeline Strategy}\label{method_explanation}
This section provides a detailed introduction of the single-prompt pipeline strategy which simplifies the LinkAlign framework for better efficiency. 

\paratitle{Align Semantics by Query Rewriting.}We propose a query semantic enhancement module that utilizes few-shot Chain of Thought examples to guide the LLMs to clarify the query's semantic intent through four reasoning steps.

\textbf{Step 1: Requirement Understanding.} The first step involves rephrasing the user query to explicitly define its objective and data requirements.

\textbf{Step 2: Key Entity Identification.} This step extracts and identifies the key entities or values from the query that are semantically relevant to the target database.

\textbf{Step 3: Entity Classification.} Based on the previous step's extractions, entities are classified into broader categories, and their relationships are defined.

\textbf{Step 4: Database Schema Inference.} The final step infers the relevant tables and columns in the target database schema that are likely to provide the necessary data to address the query.

\paratitle{Reduce Noise by Response Filtering.} We design a prompt with few-shot Chain of Thought examples to guide the LLM through multiple reasoning steps, mapping the query to the correct target database. First, the model rephrases the data requirements to ensure full understanding. Next, it evaluates each database to confirm it contains the necessary data columns. Finally, the model outputs the name of the most relevant database.

\paratitle{Identify Tables and Columns by Schema Parsing.} We adapted the prompt design from DIN-SQL, employing a single LLM call for schema linking. DIN-SQL uses a prompt with few-shot Chain of Thought examples, incorporating 10 randomly selected samples from the Spider dataset's training set. Experimental results demonstrate that this approach strikes a balance between accuracy and efficiency in database query scenarios with a small number of tables and columns, aligning with the objectives of the pipeline method. By utilizing a single model call, the method achieves strong performance in solving simpler problems.

\section{Effectiveness in Single-Database Setting}\label{sec:single_db}
In this section, we discuss how to effectively apply the LinkAlign framework to Text-to-SQL models in large-scale single-database scenarios. As a common assumption in existing studies, the single database setting simplifies real-world multi-database environments by only utilizing the schema from target database as input for Text-to-SQL models. However, the single-database setting is often impractical in real-world scenarios, as non-technical business users always struggle to select the appropriate database for data query (e.g., local SQLite and cloud BigQuery) due to a lack of expertise in database architecture. This is why our study prioritizes the multi-database setting, as this gap limits the real-world application of existing models. 

To ensure LinkAlign remains effective in the single-database setting, we need to adjust the objective of Step two to filter out irrelevant database schemas rather than those from unrelated databases. However, we notice that this approach sometimes mistakenly excludes the ground-truth schema, as it may appear unrelated to generating the correct SQL. We propose two feasible optimization techniques to address this issue as explained below. Specific adjustments of prompts and codes are available in our open source repository.

\paratitle{Random preservation with exponential decay.}
To avoid discarding potentially correct schemas without sufficient evidence, we randomly retain database schemas for each retrieval round using a dynamic retention rate. Specifically, the retention rate decays exponentially with the retrieval rounds because rewritten queries may gradually deviate from the user's original intent, thereby increasing noise. Random sampling of retrieval results not only preserves expected benefits, but also enhances the method’s generalization ability. In addition, the retention rate needs carefully chosen to ensure that the number of retained schemas is smaller than the excluded ones. In our experiments, we set the initial retention rate (at turn n = 0) to 0.55, the exponential decay coefficient to 0.6, and clip it to 0 when it falls below 1.

\paratitle{Post Retrieval.} 
This method is also highly effective by performing mini-batch retrieval on excluded database schemas (i.e., those not retrieved or filtered out). Intuitively, this offers the model a new opportunity to sift gold from the sand without the influence of the spotlight, as it compares against noisy database schemas rather than those obviously relevant. This stage employs the same method as Step One, differing only in the mini-batch retrieval scale and the number of retrieval rounds. In our experiments, we set the post-retrieval top-K to 5 and turn-n to 1.

\section{AmbiDB Dataset Construction}\label{AmbiDB_construction}
We introduce the AmbiDB benchmark, a variant of the Spider dataset. It is constructed through database expansion and query modification to better simulate real-world query scenarios characterized by large-scale synonymous databases and enhanced-ambiguity queries by multi-database contexts. The motivation stems from three limitations in existing benchmarks. First, experiments on the existing benchmarks specifies the target database required by the query in advance, ignoring the challenge of mapping user queries to the target database in multi-database scenarios. Second, the existing benchmark has a limited number of synonym databases, exhibiting lesser ambiguity in multiple databases context. Third, existing benchmarks often fail to balance database scale and variety. AmbiDB outperforms Spider in terms of scale and surpasses Bird in terms of quantity.

\subsection{Data Construction}

\paratitle{Database Expansion} We extend the database schema through two key steps. First, we instruct the LLMs to extract a subset of schema from original database.The extracted schema subset forms the foundation to construct synonym databases. These schema subsets typically capture key characteristics of the specific domain. For example, the Student table, containing attributes such as student ID and name, can serve as a shared schema for synonym databases related to College domain. Second, we add new tables and  columns to expand the scale of every database while aligned with the original database themes. The expanded schemas preserve the integrity of the original SQL queries while becoming larger in scale. Furthermore, the inclusion of similar sub-schemas across multiple databases enhances contextual ambiguity, making it more close to real-world query scenarios.

\paratitle{Query Modification } We instruct the LLMs to modify original queries with the use of synonymous databases, which making them more complex and ambiguous in multi-database scenarios. The motivation for this step is that the original queries are semantically explicit or contain sufficient details, facilitating identifying required tables and columns from the target database. 
This step aligns the queries semantically with overlapping schemas included in multiple synonymous databases, while avoiding tokens with identical names across databases. Additionally, we subtly incorporated details that could be reasoned to help determine the target database, preventing complete confusion. 

\subsection{Data Filter}
In the main text, we provided a brief overview of Query Modification. However, the complete process involves two key steps: (1) generating the correct SQL query for a given question, and (2) filtering the rewritten question-SQL pairs.

\paratitle{Generating the Correct SQL Query for the Question.} LLMs with strong reasoning and generation capabilities is employed to generate the corresponding SQL query based on the modified question. Due to the inherent ambiguity and complexity of the questions, the model cannot guarantee absolute correctness of the generated SQL. To address this, we implement an automatic verification step, where the model checks if the data requirements described in the question align with the SQL query’s execution result. The model is then given a single opportunity to correct any errors by fine-tuning the question to match the SQL query. Finally, we manually review and verify the question-SQL pairs to ensure correctness.

\paratitle{Filtering the Rewritten Question-SQL Pairs.} We assess the complexity and ambiguity of each question individually, removing or modifying any inadequate samples. We then filter out the top 10\% and bottom 5\% of questions based on length. This approach is driven by two considerations: first, longer questions may contain more semantically relevant information about the target database, making it easier to locate the question through semantic matching alone, which fails to simulate the real-world challenges of database localization. Second, shorter questions may oversimplify the Schema Linking task, making it easier to address the second challenge.

 \section{Supplementary Experimental Setup}\label{comparable_benchmark}
 \paratitle{SPIDER} includes 10,181 questions and 5,693 SQL queries spanning 200 databases, encompassing 138 domains. The dataset is split into 8,659 training examples, 1,034 development examples, and 2,147 test examples. The databases utilized in the multi-database scenario experiment are primarily sourced from the dev dataset and the train dataset.

\paratitle{BIRD} comprises 12,751 question-SQL pairs across 95 large databases and spans 37 professional domains. It adds external knowledge to align the query with specific database schemas. The queries in BIRD are more complex than SPIDER, necessitating the provision of sample rows and external knowledge hints to facilitate schema linking.

\paratitle{SPIDER 2.0} comprises 632 real-world text-to-SQL tasks from enterprise databases with thousands of columns and diverse SQL dialects. It requires schema linking, external documentation, and multi-step SQL generation, posing greater challenges than SPIDER 1.0 and BIRD. Simplified variants, Spider 2.0-lite and Spider 2.0-snow, focus on text-to-SQL parsing without workflow interaction.

\section{End-to-End Efficiency Analysis}
To further assess LinkAlign's efficiency, we conducted an additional evaluation on the Spider 2.0-Lite benchmark, which is recognized for its large-scale and realistic database settings. We randomly sampled 50 examples and adopted DeepSeek-V3 as the backbone to measure the end-to-end SQL generation efficiency of different baseline methods. As reported in Table~\ref{tab:efficiency}, LinkAlign consistently improves the quality of SQL generation, without incurring notable overhead in runtime or token usage. In particular, the Pipeline mode exhibits clear efficiency gains over baselines that directly process the full schema (e.g., DIN-SQL), highlighting its effectiveness in balancing performance and cost.

\begin{table}[t]
\centering
\small
\begin{tabular}{lcc}
\hline
\textbf{Methods} & \textbf{Avg. Time (s)} & \textbf{Avg. Token} \\
\hline
LinkAlign-Pipeline & 127.6 & 8507.4 \\
LinkAlign-Agent    & 183.5 & 12486.5 \\
CHESS                & 457.8 & 21413.8 \\
RSL-SQL              & 157.2 & 14713.4 \\
DIN-SQL              & 146.3 & 8600.0 \\
\hline
\end{tabular}
\caption{End-to-end run time efficiency comparison. LinkAlign achieves improved schema linking without significant increases in runtime or token consumption.}
\label{tab:efficiency}
\end{table}

\section{Limitations}
This paper presents two limitations that will require attention in future work. First, we did not explore the potential advantages of combining different strategies for addressing the schema linking task, even though our modular interface could facilitate such combinations. Second, we did not use the most advanced LLMs in our experiments. The reasoning limitations of the LLMs maybe resulte in more noticeable performance between different methods. However,as large models gain sufficient capabilities, we may need to consider whether it's necessary to simplify the framework.

\end{document}